\documentclass[authoryear,preprint,11pt,pdf]{elsarticle}
\usepackage[utf8]{inputenc}
\usepackage[TU]{fontenc}
\usepackage{natbib}
\usepackage{booktabs}
\usepackage{graphics}
\usepackage{pdflscape}
\usepackage{amsmath}
\usepackage{amsfonts}
\usepackage{amssymb}
\usepackage{lmodern}
\usepackage{bm}
\usepackage{caption}
\usepackage{subcaption}
\begin{document}
\begin{frontmatter}
  
\title{Multicriteria interpretability driven Deep Learning}

\author[add1,add2]{Marco Repetto}
\ead{marco.repetto@unimib.it}
\address[add1]{Siemens Italy, Digital Indutries, Milan, Italy}
\address[add2]{University of Milano-Bicocca, Department of Economics, Management and Statistics, Milan, Italy}
\date{\today}

\begin{abstract}
Deep Learning methods are renowned for their performances, yet their lack of interpretability prevents them from high-stakes contexts.
Recent model agnostic methods address this problem by providing post-hoc interpretability methods by reverse-engineering the model's inner workings.
However, in many regulated fields, interpretability should be kept in mind from the start, which means that post-hoc methods are valid only as a sanity check after model training.
Interpretability from the start, in an abstract setting, means posing a set of soft constraints on the model's behavior by injecting knowledge and annihilating possible biases.
We propose a Multicriteria technique that allows to control the feature effects on the model's outcome by injecting knowledge in the objective function.
We then extend the technique by including a non-linear knowledge function to account for more complex effects and local lack of knowledge.
The result is a Deep Learning model that embodies interpretability from the start and aligns with the recent regulations.
A practical empirical example based on credit risk, suggests that our approach creates performant yet robust models capable of overcoming biases derived from data scarcity.
\end{abstract}

\end{frontmatter}

\section{Introduction}
Deep Learning (DL) models are used extensively nowadays in many fields ranging from self-driving cars \citep{rao2018} to brain-computer interfaces \citep{zhang2019} to gaming \citep{vinyals2019}.
Recent software and hardware democratized DL methods allowing scholars and practitioners to apply them in their fields.
On the software side, recent frameworks as Tensorflow \citep{abadi2015} and PyTorch \citep{paszke2019} allowed to create complex DL models avoiding the need to write ad-hoc compilers as did by \cite{lecun1990}.
On the hardware side, the decrease in the cost of the necessary hardware to train such models, allowed many people to build and deploy sophisticated Neural Networks with minimal costs \citep{zhang2018}.
The democratization of such powerful technologies allowed many fields to benefit from it aside from computer science.
Some of those that benefitted the most are Economics \citep{nosratabadi2020}, and Finance \citep{ozbayoglu2020}.
DL applications have piqued the interest of governments, who are concerned about possible social implications.
It is well known that these models necessitate extra vigilance when it comes to training data in order to minimize biases of any kind, especially in high-stakes judgments \citep{rudin2019a}.
To counter these side effects, the governments enacted several regulatory standards, and the jurisprudence started to elaborate on the right to explanation concept \citep{dexe2020}.
In this effort to build interpretable but DL grounded models, scholars have started developing post-hoc interpretation methods.
These approaches, however, are at odds with what is prescribed by recent guidelines, requiring interpretability from the start \citep{EC2019}.
Another issue is that such approaches focus only on the interpretation after a model's training and cannot be used to insert prior information or remove biases.

This work focuses on ensuring the interpretability of DL models from the beginning through knowledge injection and investigating their potential in empirical settings as in credit risk prediction.
In this regard, we make three contributions to the literature.
First, we allow the Decision Maker (DM) to inject previous knowledge and alleviate dataset biases in the model training process by controlling the features' effect. 
Physics-guided Neural Networks (PGNN) are based on similar methodologies that are widely employed in physics-related applications \citep{daw2021}.
Knowledge injection, whether physics related or not, generally involves some sort of constraints in terms of features relationship with the outcome that can be implemented as posed by \cite{vonrueden2021} in four different ways: (i) on the training data; (ii) on the hypothesis set; (iii) on the learning algorithm; (iv) on the final hypothesis. 
Our methodology infuses knowledge on the learning algorithm level as this procedure relates to some post-hoc interpretability methods. According to our understanding, these architectures were never proposed outside of the physics and engineering fields. 
And even in such areas, effects constraints were conditional on the context as in \cite{muralidhar2018} or applied to non-DL techniques \citep{kotlowski2009, lauer2008, vonkurnatowski2021a}. 
Our approach applies to any DL architecture and is not conditional on features' context. 
We test the validity of our technique in credit risk as the concept of Sustainable AI will considerably impact this field. 
The recent frameworks, as the one proposed by \cite{bucker2021} do not allow for interpretability from the start as posed by \cite{EC2019}. 
These techniques can spot biases but cannot counter them, as their scope is only explainability and not knowledge injection. 
Our methodology can handle both these aspects by leaving a model that is compliant with the new guidelines on Sustainable AI.
Second, we allow for non-linear effects and local lack of knowledge by defining ad-hoc knowledge functions on models parameters.
This additional specification is necessary for two reasons. 
To begin, the empirical literature in credit risk agrees that the performances of DL models are due primarily to the fact that they capture non-linear patterns \citep{ciampi2013small}.
The second reason for including this non-linear pattern is that knowledge may be lacking in some regions of the feature space.
Third, we explore the relationship between post-hoc interpretability methods that are model agnostic, as for the case of Accumulated Local Effects \citep{apley2020visualizing}.
These methods play two critical roles in our strategy.
Initially, they provide the DM with graphical visualizations that allow him to communicate with non-experts.
Second, they serve as sanity checks for our methodology and hyperparameter optimization based on explainability.

The rest of this paper is structured as follows.
The knowledge injection in the model and the multicriteria problem development are all covered in Section 2.
The data sample utilized to test our technique, software packages, and hardware is briefly discussed in Section 3.
Section 4 summarizes the findings and examines the most important ones.
Section 5 comes to a close.

\section{Methodology}
\subsection{Deep Learning}
DL is an AI subfield and type of Machine Learning technique aimed at developing systems that can operate in complex environments \citep{goodfellow2016}. Deep architectures underpin DL systems \citep{bottou2007} which can be defined as: 

\begin{equation}
\mathcal{F} = \{f(\cdot, w), w \in \mathcal{W} \}
\end{equation}

where $f(\cdot, w)$ is a shallow architecture as for example the Perceptron proposed by \cite{rosenblatt1958}.
Prior to Rosenblatt, \cite{mcculloch1943} proposed a system in which binary neurons arranged together could perform simple logic operations.
Nowadays, neither the Perceptron nor the system proposed by \cite{mcculloch1943} is used in current Artificial Neural Networks (ANNs) configurations.
Modern architectures rely on gradient-based optimization techniques and particularly on Stochastic Gradient Descent (SGD) \citep{saad1998}.
One of these first architectures trained through gradient-based methods was the Multilayer Perceptron (MLP) \citep{rumelhart1986} followed by the Convolutional Neural Network proposed by \cite{lecun1990}.

This paper tests our approach on two DL architectures: the MLP and the ResNet (RN).
The choice of using the MLP is because it is somehow the off the shelve solution in many use-cases and especially in credit risk \citep{ciampi2021rethinking} and in several works on retail credit risk \citep{lessmann2015}.
MLP consists of a direct acyclic network of nodes organized in densely connected layers. After being weighted and shifted by a bias term, inputs are fed into the node's activation function and influence each subsequent layer until the final output layer.

In a binary classification task, the output of an MLP can be described as in \cite{arifovic2001} by:

\begin{equation}
	f(x) = \phi \left(\beta_0 + \sum_{j=1}^d\beta_jG \left(\gamma_{j0}+\sum_{i=1}^p \gamma_{ji}x_i \right)\right)
\end{equation}

\begin{figure}[h]
	\begin{center}
		\begin{tabular}{ccc}
			
			\includegraphics{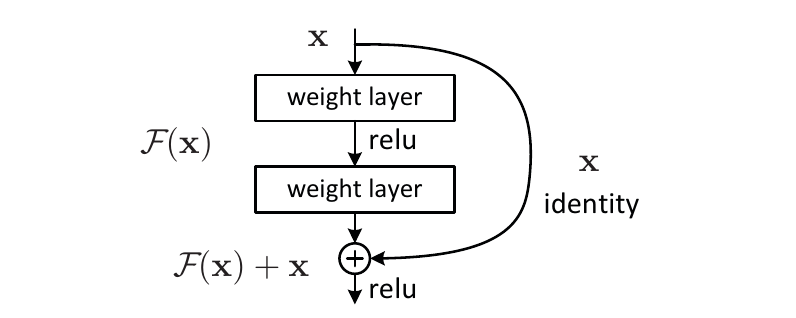}
			
		\end{tabular}
              \end{center}
	\caption{Residual Network skip connection block.}
	\label{resnet}
\end{figure}

RN differs from the canonical MLP architecture since it has "shortcut connections" that mitigate the problem of degradation in the case of multiple layers \citep{he2015}.
Although the usage of shortcut connections is not new in the literature \citep{venables1999}, the key proposal of \cite{he2015} was to use identity mapping instead of any other nonlinear transformation.
Figure \ref{resnet} shows the smallest building block of the ResNet architecture in which both the first and the second layers are shortcutted, and the inputs $x$ are added to the output of the second layer.

\subsection{Multicriteria optimization}
Multiple Criteria Decision-Making (MCDM) is a branch of Operations Research and Management Science.
MCDM represents the set of methods and processes through which the concern for several conflicting criteria can be explicitly incorporated into the analytical process \cite{ehrgott2002}.
Several strategies, spanning from a priori to interactive ones, have been developed to address this issue \citep{miettinen2012nonlinear}. 
Historically MCDM's first roots are the ones laid by Pareto at the end of the 19th century. 
However, the modern MCDM has its origins more recently in the work of \cite{koopmans1951} who modified the canonical single objective linear programming model by reframing it as a vector optimization problem.
An MCDM problem takes the following form:

\begin{alignat}{3}
&\min_{\pmb{x}}     &\qquad& \{f_1(\pmb{x}), f_2(\pmb{x}), \dots, f_k(\pmb{x})\}\label{eq:optProb}\\
&\text{subject to} &      & \pmb{x} \in S\label{eq:constraint1}
\end{alignat}  

where $f_i: {\rm I\!R}^n \rightarrow {\rm I\!R}$ is the ith objective and the vector $\pmb{x}$ contains the decision variables that belong to the feasible set $S$.
Scalarization is a common approach for dealing with Multicriteria optimization problems.
By scalarization, a vector optimization problem turns into a single objective optimization problem.
Additional concerns affect the DM's preferences scheme as a result of this.
In our approach, we begin by using a weighted sum scalarization to handle our problem:

\begin{alignat}{3}
&\min_{\pmb{x}}     &\qquad& \pmb{w}^\top\pmb{f}(\pmb{x}) \label{eq:optProb}\\
&\text{subject to} &      & \pmb{x} \in S\label{eq:constraint1}
\end{alignat} 

where the weights are the relative preference of the DM toward a specific goal.
The incorporation of such preferences can happen in two ways, either a priori or a posteriori.
In our approach, we use an a posteriori method as this best suits the DM's lack of knowledge, which may be uncertain about the relative importance of each objective.

\subsection{Knowledge injection}
Knowledge in this paper is intended as validated information about relations between entities in specific contexts as in \cite{vonrueden2021}. 
The key feature of such a definition is that it allows for formalization, implying that such knowledge can be transformed into mathematical constraints somehow.

Let's assume we have a deep architecture such that $\hat{y} = \mathcal{F}(\pmb{x}, \mathcal{W})$ and we observe the true label $y$ then we can train a supervised model by using a differentiable loss function as for the case of regression by using the mean squared error (MSE):

\begin{equation}
Loss_f(\pmb{x}, \mathcal{W}) = \frac{1}{n} \sum_{i=1}^n(y_i - \hat{y}_i)^2
\end{equation}

or in our case, of a binary classification with the binary cross-entropy:

\begin{equation}
Loss_f(\pmb{x}, \mathcal{W}) = \frac{1}{n} \sum_{i=1}^n\left[ y_i \log(\hat{y}_i + (1-y_i) \log (1-\hat{y}_i)\right]
\end{equation}

this will constitute the first objective of our Multicriteria loss function, that is, data fitting.
The knowledge injection instead will act on the features' effects on the model outcome, which means that our knowledge-based objective will be:

\begin{equation}
Loss_k(\pmb{x}, \mathcal{W}) = k(\pmb{x}) \odot \frac{\delta \mathcal{F}(\pmb{x}, \mathcal{W})}{\delta \pmb{x}} 
\end{equation}

where the right-hand side of the Hadamard product is the gradient of our DL model at the feature level $\pmb{x}$ whereas $k(\pmb{x})$ is a function that penalizes/favorites specific effect of the gradient with range $[-1,1]$.
Since knowledge is hardly spread through the whole feature space, a possible strategy is to define $k(\pmb{x})$ that maps the feature space to the knowledge we expect on that particular feature neighbor.

In its most straightforward formulation, $k(\pmb{x})$ can be a scalar, and in this formulation is easier to investigate its behavior at the model level.
Let us assume $k(\pmb{x}) = \pmb{1}$ then what is enforced is that all partial derivatives should be negative therefore enforcing monotonicity and, in particular decreasing monotonicity.
The opposite holds for the case when $k(\pmb{x}) = \pmb{-1}$.
When $k(\pmb{x}) = \pmb{0}$ there is no constraint on the gradient behavior, meaning that knowledge is non-existent and therefore not injected.
Following \cite{daw2021} we can augment our Multicriteria function with a further constraint that measures the network complexity as, for example, an $L_2$ regularization on the weights.
The result is the following unconstrained minimization problem:

\begin{equation}
\min_{\mathcal{W}} \pmb{\lambda}^\top \begin{bmatrix} Loss_f(\pmb{x}, \mathcal{W})  \\ ||\mathcal{W}||_2 \\ Loss_k(\pmb{x}, \mathcal{W})  \end{bmatrix}
\end{equation}  

\subsection{Interpretability methods}
Model interpretability is gaining popularity due to the increasing applications of non-linear models \citep{molnar2020}.
Model-agnostic and model-aware approaches are the two main types of interpretability approaches.
The model is never accessed in model-agnostic interpretability methods.
Unlike the model-agnostic approach, the model-aware technique has access to model parameters such as the gradient.
Model-agnostic methods apply to all models, whereas model-aware are restricted to a specific class of models.
On the other hand, model-aware approaches are more efficient and converge faster to the genuine interpretability metric.

\subsubsection{Model-aware interpretability}
One of the first techniques proposed by the literature was to use the product of the model's gradient with feature values \citep{baehrens2010explain}.
\cite{simonyan2014} proposed Saliency Maps based on the gradient of model output with respect to the input features. 
In other words, the score of multiclass $c$ classifier can be locally approximated by:

\begin{equation}
S_c(I) \approx w_c^T I + b_c
\end{equation}

where $w_c^T$ is the gradient of the model at that particular class $c$ in that particular feature configuration $I_0$:

\begin{equation}
w = \frac{\partial S_c}{\partial I} \Bigr|_{I_0}
\end{equation}

\cite{sundararajan2017} through an axiomatic approach, questioned the validity of such procedure, arguing that using only the gradient may result in misleading feature attributions with respect to a baseline observation.
They proposed the concept of Integrated Gradients, a path-dependent approach in which the gradients are accumulated over the linear combinations between the observation and a baseline.  
This implies the evaluation of the following:
\begin{equation}
IG_0 = (I_0- I') \int^1_{\alpha=0} \frac{\partial S_c(I' + \alpha (I_0 - I')}{\partial I} d \alpha
\end{equation}

\subsubsection{Model-agnostic interpretability}
One of the first model-agnostic technique was the Partial Dependence (PD) proposed by \cite{friedman1991}.
PD plots evaluate the change in the average predicted value, as specified features vary over their marginal distribution \cite{goldstein2015}.
However the main limitation of the PD is the dependence within features since evaluating the PD carries the risk of computing points outside the data envelope.
\cite{apley2020visualizing} proposed the Accumulated Local Effects (ALEs) to address this flaw.
ALEs have the advantage of avoiding the problem of evaluating variables' effects outside the data envelope \citep{apley2020visualizing}.
Computing the ALE implies the evaluation of the following:

\begin{equation}
\label{ALE_est}
ALE_{\hat{f}, S} (x_S) = \sum_{k=1}^{k_S(x)}\frac{1}{n_S(k)} \sum_{i:x_S^{(i)} \in N_S (k)} \left[\hat{f}(z_{k,j}, x_{\setminus S}^{(i)}) - \hat{f}(z_{k-1,j}, x_{\setminus S}^{(i)}) \right] - C
\end{equation}

where: $\hat{f}$ is the black-box model, $S$ is the subset of variables' index, $X$ is the matrix containing all the variables, $x$ is the vector containing the feature values per observation, $z$ identifies the boundaries of the K partitions, such that $z_{0,S} = min(x_S)$, $C$ is a constant term to center the plot.
To make Equation \ref{ALE_est} model-aware we can substitute finite differences with the gradient as $\frac{\delta \hat{f}(z_S, X_{ \setminus S})}{\delta z_S} \approx \hat{f}(z_{k,j}, x_{\setminus S}^{(i)}) - \hat{f}(z_{k-1,j}, x_{\setminus S}^{(i)}) $.
Because of that, the resulting model-aware formula is:

\begin{equation}
\label{ALE_eq}
ALE_{\hat{f}, S} (x) = \int_{z_{0,S}}^x \left[ \int \frac{\delta \hat{f}(z_S, X_{ \setminus S})}{\delta z_S} d\mathcal{P}(X_{\setminus S}|z_S)\right] dz_S - C
\end{equation}

As a result, knowledge injection, as proposed in the previous section, will have an effect on the ALEs because the final model will have a different gradient than the non-knowledge injected one.

\section{Data, software, hardware}
\subsection{Data}
To test the goodness of our approach, we provide an empirical application in the context of credit risk.
In particular on the problem of bankruptcy prediction.
We used a publicly available dataset of Polish enterprises donated to the UCI Machine Learning Repository by \cite{zikeba2016ensemble}.
The data contains information about the financial conditions of Polish companies belonging to the manufacturing sector.
The dataset contains 64 financial ratios ranging from liquidity to leverage measures \footnote{For a complete description of the financial indicators, please consider Table \ref{tab:app1} in the Appendix.}.
Moreover, the dataset distinguishes five classification cases that depend on the forecasting period.
In our empirical setting, we focus on bankruptcy status after one year.
In this subset of data, the total number of observations is 5910, out of which only 410 represents bankrupted firms.
It is worth noting that we do not counter the class imbalance in the empirical setting, although this is something done commonly in the literature.
We retained class imbalance to test the robustness of our approach even in conditions of scarcity of a particular class and used robust metrics such as the Area Under the Receiving Operating Curve (AUROC).
Moreover, as our empirical experiment focuses on testing our approach on model interpretability, we restricted the number of features we considered to six.
This is due to the fact that ALEs are inspected as plots, and having a plot for each feature increases complexity without providing any additional benefit to the reader or our approach.
The choice was to focus on Attr 13, Attr 16, Attr 23, Attr 24, Attr 26, and Attr 27.
The attributes were selected by using a ROC-based feature selection \citep{kuhn2019feature}.

\subsection{Software and hardware}
The overall pipeline management is built in R \citep{rcoreteam2020}.
The preprocessing relied on the tidymodels ecosystem \citep{kuhn2020a} as well as on the tidyverse \citep{wickham2019}.
The DL models are developed in Julia \citep{bezanson2017julia} using the Flux framework \citep{Flux.jl-2018, innes:2018}.
The interoperability between the two languages is possible via the JuliaConnectoR library \citep{lenz2021}.
To debug the model and to check the validity of our approach, we employed the ALEPlot package \citep{apley2018}.
As for the hardware environment, the pipeline is carried out on a local machine with 12 logical cores (Intel i7-9850H), 16 GB RAM, and a Cuda enabled graphic card (NVIDIA Quadro T2000).
Both Julia and R codes are freely available for research reproducibility on \href{https://gitlab.com/mrepetto94/multicriteria-interpretability-driven-deep-learning}{GitLab}, and an ad-hoc Docker container has been created on \href{https://hub.docker.com/repository/docker/mrepetto94/multicriteria-interpretability-driven-deep-learning}{DockerHub}.

\section{Results}
To test the performance of our approach on the Polish firms' dataset, we use a standard practice in the field of DL. 
At first, we split the dataset into training and testing.
Three-quarters of the dataset is for training the model and the rest for testing its performance. 
In the case of a model that contains no hyperparameters, a setting like this will suffice. 
However, in DL, this is never the case as these models require a thorough calibration of hyperparameters.
In our setup, the hyperparameters are the elements contained in $\pmb{\lambda}$.
Therefore a common strategy is to use the training set to do what is called hyperparameter optimization \citep{goodfellow2016}.
Therefore, the training set is further divided into training and validation sets, and the model is fitted and validated with different parameters.
In our case, we draw from the training set ten samples using the bootstrap technique as proposed by \cite{efron1997}, and we trained our model using different combinations of hyperparameters.
As for the choice on implementing such hyperparameter search, we relied on grid search, also known as full factorial design \citep{montgomery2017design}.
We then trained the model on the entire training set and classified bankruptcy state on the test set with the optimal hyperparameters.

For a deep understanding of the results, we divided the analysis into three parts. 
Firstly we performed hyperparameter optimization and subsequent hold-out testing using both the MLP and the RN, with the former being the best performing model.
Secondly, with the MLP, we analyzed the effect on the interpretation by using ALE plots.
Thirdly we tested the robustness of our approach by diminishing the amount of data.

\subsection{Performance review}
Table \ref{tab:table1} presents the mean AUROC as well as its standard errors as our model validation involved ten bootstrap samples.
The first striking result is that both the MLP and RN perform poorly without any regularization nor knowledge injection.
This result is in line with \cite{zikeba2016ensemble}, which also found that the ANN architectures suffer in terms of generalization.
What instead is of great interest is the increase in performance when both  
regularisation and knowledge injection are brought to the table.
In particular, the ResNet seems to perform better with low levels of knowledge injection and a modest level of regularization.
However, what is more, relevant is the behavior of the MLP.
The model is more sensitive to knowledge injection and performs significantly better in model validation. 
This result is evident when $\lambda_3 = 0.3$, 
all the MLP models are above an AUROC of $0.8$ with a modest amount of standard error.
Another significant result of the MLP is performance deterioration when regularization starts to ramp $\lambda_2 = 0.3$ up even with injected knowledge.
What is worth noting is that the dataset suffers from significant class imbalance, and nothing has been done to alleviate it to test the effectiveness of our approach in such context.
Indeed knowledge injection alleviates misclassification and creates a model that is on par with other robust classifiers.

\begin{table}[]
  \centering
  \caption{Performances of Multilayer Perceptron and Residual Network on the training set with various hyperparameter settings. The performance is measured as the mean and standard errors of the Area Under the Receiving Operating Curve in each bootstrap sample. Bold values indicate the best performing hyperparametrization for each model.}
  \begin{tabular}{@{}llrlrrlrr@{}}
\toprule
\multicolumn{1}{c}{\textbf{}} & \multicolumn{1}{c}{\textbf{}} & \multicolumn{1}{c}{\textbf{}} &  & \multicolumn{2}{c}{\textbf{Multilayer Perceptron}} &  & \multicolumn{2}{c}{\textbf{Residual Network}} \\ \cmidrule(lr){5-6} \cmidrule(l){8-9} 
$\lambda_1$ & $\lambda_2$ & $\lambda_3$ &  & Mean & Standard error &  & Mean & Standard error \\ \midrule
1.0 & 0.0 & 0.0 &  & 0.6585 & 0.0178 &  & 0.5061 & 0.0843 \\
0.9 & 0.1 & 0.0 &  & 0.6924 & 0.0111 &  & 0.5308 & 0.0805 \\
0.8 & 0.2 & 0.0 &  & 0.6302 & 0.0757 &  & 0.6326 & 0.0180 \\
0.7 & 0.3 & 0.0 &  & 0.7175 & 0.0059 &  & 0.5584 & 0.0900 \\
0.9 & 0.0 & 0.1 &  & 0.7905 & 0.0087 &  & 0.6418 & 0.0740 \\
0.8 & 0.1 & 0.1 &  & 0.8286 & 0.0149 &  & 0.5744 & 0.0769 \\
0.7 & 0.2 & 0.1 &  & 0.7586 & 0.0336 &  & 0.5059 & 0.0746 \\
0.6 & 0.3 & 0.1 &  & 0.6163 & 0.1219 &  & \textbf{0.6604} & 0.0879 \\
0.8 & 0.0 & 0.2 &  & 0.8263 & 0.0129 &  & 0.5124 & 0.0664 \\
0.7 & 0.1 & 0.2 &  & 0.8242 & 0.0170 &  & 0.6102 & 0.0186 \\
0.6 & 0.2 & 0.2 &  & 0.8206 & 0.0123 &  & 0.5037 & 0.0443 \\
0.5 & 0.3 & 0.2 &  & 0.7617 & 0.0601 &  & 0.6249 & 0.0593 \\
0.7 & 0.0 & 0.3 &  & 0.8202 & 0.0119 &  & 0.6074 & 0.0172 \\
0.6 & 0.1 & 0.3 &  & 0.8289 & 0.0139 &  & 0.5410 & 0.0417 \\
0.5 & 0.2 & 0.3 &  & \textbf{0.8306} & 0.0135 &  & 0.5318 & 0.0150 \\
0.4 & 0.3 & 0.3 &  & 0.8178 & 0.0198 &  & 0.5628 & 0.0378 \\ \bottomrule
\end{tabular}
\label{tab:table1}
\end{table}

The performances in Table \ref{tab:table1} are promising. 
However, to precisely measure model generalization error, the performances that need to be considered are those taken from the test set.
Table \ref{tab:table2} presents these performances by taking into account only the optimally parametrized models and their baseline, that is, the model with $\lambda_1 = 1$.
The clear-cut result from this table is that the MLP generalizes way better than its counterpart, with a slight decrease in performance in line with the expectations. 
This result suggests that knowledge injection corroborated with mild regularization can enhance the generalization performances of a DL classifier as the MLP and make it robust to class imbalances.

\begin{table}[]
  \centering
  \caption{Performances of Multilayer Perceptron and Residual Network on the test set with validated and baseline hyperparameter settings. The performance is measured as the Area Under the Receiving Operating Curve in the test sample.}
\begin{tabular}{@{}lllrr@{}}
\toprule
$\lambda_1$ & $\lambda_2$ & $\lambda_3$ & \multicolumn{1}{c}{\textbf{Multilayer Perceptron}} & \multicolumn{1}{c}{\textbf{Residual Network}} \\ \midrule
1.0 & 0.0 & 0.0 & 0.577 & \textbf{0.582} \\
0.5 & 0.2 & 0.3 & \textbf{0.821} & - \\
0.4 & 0.3 & 0.1 & - & 0.518 \\ \bottomrule
\end{tabular}
\label{tab:table2}
\end{table}

In the following sections, we will investigate further the performance of the MLP with and without knowledge injection in terms of interpretability and robustness to data scarcity.
We will focus only on the MLP, as it was the most performant architecture.
Moreover, analyzing the interpretations of a non-performant classifier as the RN has no useful meaning from a practical point of view.

\subsection{Interpretability review}
\begin{figure}
     \centering
     \begin{subfigure}[b]{\textwidth}
         \centering
         \includegraphics[width=\textwidth]{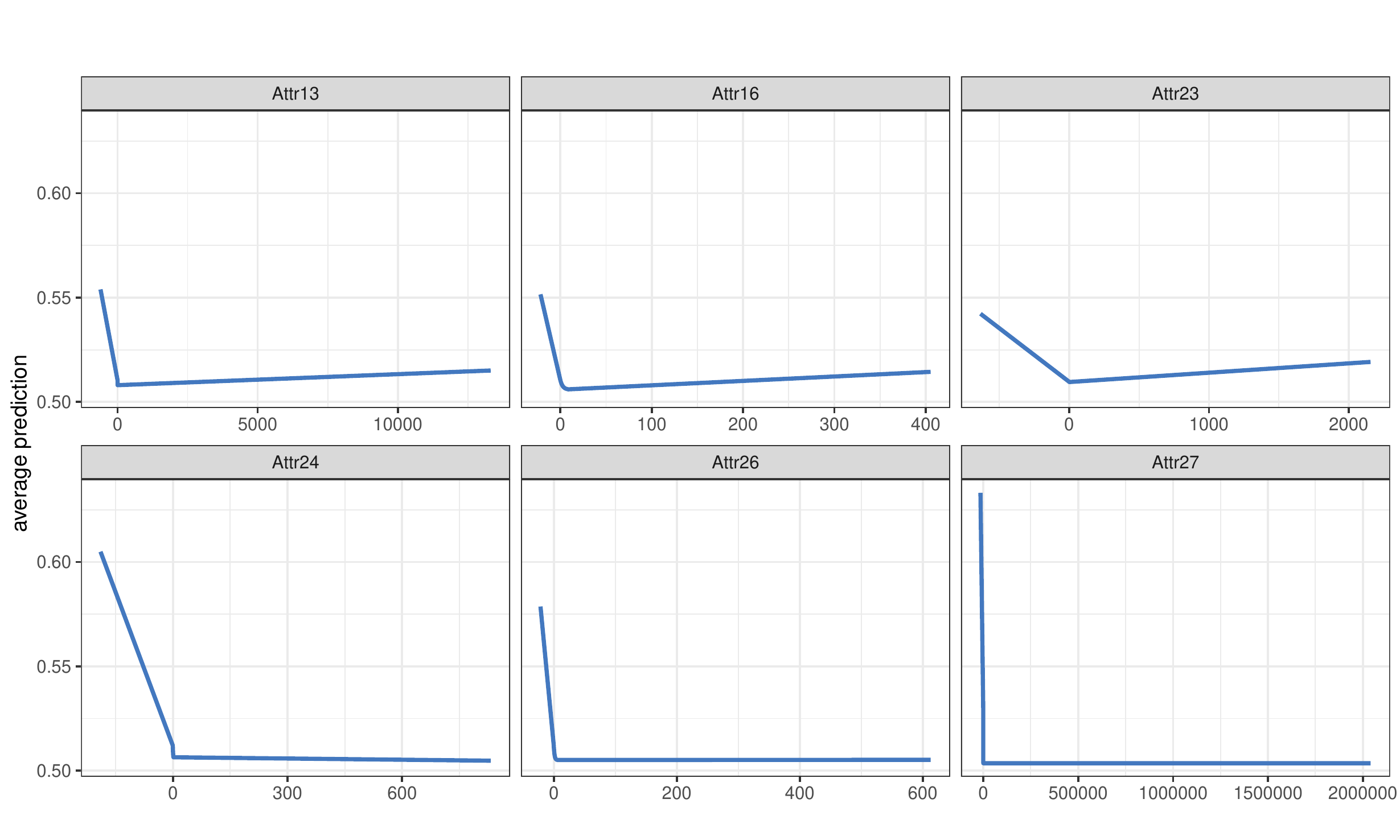}
         \caption{Accumulated Local Effects plot of the Multilayer Perceptron architecture, without regularization and knowledge injection (i.e. $\lambda_1 = 1, \lambda_2 = 0.0, \lambda_3 = 0.0$).}
         \label{fig:no_k}
     \end{subfigure}
     \hfill
     \begin{subfigure}[b]{\textwidth}
         \centering
         \includegraphics[width=\textwidth]{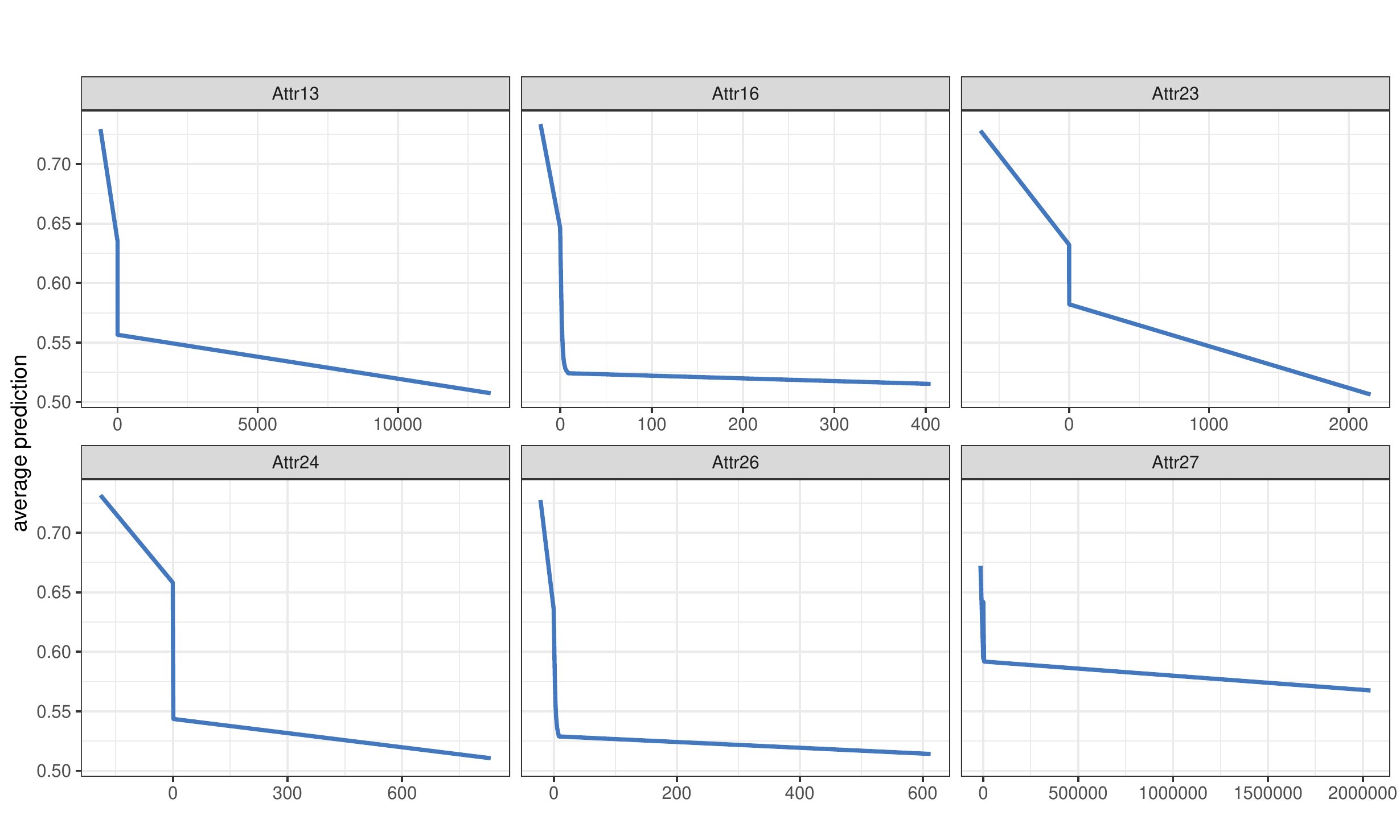}
         \caption{Accumulated Local Effects plot of the Multilayer Perceptron architecture, with regularisation and knowledge injection optimally selected from the hyperparameter optimization procedure (i.e. $\lambda_1 = 0.5, \lambda_2 = 0.2, \lambda_3 = 0.3$).}
         \label{fig:with_k}
     \end{subfigure}
        \caption{Accumulated Local Effects plot of the Multilayer Perceptron architecture, with and without regularization and knowledge injection.}
        \label{fig:ale_plot}
\end{figure}

A mentioned in the previous sections, post-hoc interpretability methods are essential tools for model debugging and to inspect any model bias.
As a result, we present figure \ref{fig:no_k}, which demonstrates MLP ALEs with and without knowledge injection.
The ALEs of the model without knowledge injection show several misbehaviors that may be due to class imbalance or hidden biases in the training sample.
In detail:

\begin{itemize}
\item Attr 13: which is also known as the EBITDA-To-Sales ratio, is a profitability indicator. Therefore we should expect to decrease the probability of bankruptcy, especially in cases where the ratio is positive. The opposite occurs instead. An increase of the ratio above zero increases the probability of bankruptcy.  This effect is at odds with the literature on the subject as, for example, in \cite{platt2002predicting}.
\item Attr 16: is the inverse and a proxy of the Debt-To-EBITDA ratio which is leverage ratio. For the inverse of a leverage ratio we would assume negative impact on bankruptcy as in \cite{beaver1968alternative}.
\item Attr 23: is the Net profit ratio and is a productivity ratio \cite{lee2013} which tends to have a negative impact on bankruptcy.
\end{itemize}

To counter this common biased effects we assumed the following logistic form for all the features' knowledge function:

\begin{equation}
  k(x) = \frac{1}{1+e^{-100x}}
\end{equation}

Such a knowledge function penalizes only positive effects above zero and retains the correctly captured effects below it.
With this setting, in the case of moderate knowledge injection, the effects align with the literature findings.

\subsection{Robustness checks}
A fundamental question is how model performances deteriorate with less training data.
In previous works, knowledge injection has been implemented indeed to alleviate such problems as in \cite{vonkurnatowski2021a}.
To discover how our approach deals with scarce data, we systematically diminished the training set and measured the corresponding performance on the test set.
These results are in Table \ref{tab:data_scarcity} which depicts the different performances in the test set as the training data proportion decreases.
In concordance with the literature on knowledge injection, our approach prevents performance degradation even in extreme cases where only half the dataset is used for training.

\begin{table}[]
  \centering
  \caption{Performances of Multilayer Perceptron on the test set under different proportions of train/test split. The performance is measured as the Area Under the Receiving Operating Curve in the test sample.}
  \begin{tabular}{lrr}
\hline
\multicolumn{1}{c}{\textbf{Train/Test}} & \multicolumn{1}{c}{\textbf{\begin{tabular}[c]{@{}c@{}}With knowledge\\ ($\lambda_1 = 0.5, \lambda_2 = 0.2, \lambda_3 = 0.3$)\end{tabular}}} & \multicolumn{1}{c}{\textbf{\begin{tabular}[c]{@{}c@{}}Without knowledge \\ ($\lambda_1 = 0.1, \lambda_2 = 0.0, \lambda_3 = 0.0$)\end{tabular}}} \\ \hline
0.85 & 0.829 & 0.615 \\
0.80 & 0.828 & 0.543 \\
0.75 & 0.821 & 0.577 \\
0.7 & 0.822 & 0.605 \\
0.65 & 0.790 & 0.613 \\
0.6 & 0.817 & 0.646 \\
0.55 & 0.803 & 0.505 \\
0.5 & 0.823 & 0.641 \\ \hline
\end{tabular}
  \label{tab:data_scarcity}
\end{table}

\section{Conclusion}
In this paper, we presented a novel approach to knowledge injection at the level of feature effects of a DL model.
Model interpretability is a particularly crucial topic, and recent legislation implies interpretability from the start.
Recent post-hoc interpretability methods fail to provide this.
Our approach solves the problem by allowing to control model interpretation from the start.
The approach consists in solving a Multicriteria minimization problem in which knowledge adherence competes with greedy data fitting and regularization.
We accounted for partial knowledge and nonlinearity by defining ad-hoc knowledge functions on the model's parameters.
We presented a use case of bankruptcy prediction using a Polish firm's dataset to test our approach.
The results suggest that knowledge injection improves performances and keeps model interpretation in line with the literature findings, avoiding idiosyncratic effects resulting from either class imbalance or possible biases in the dataset.
The DM can check the effects of our approach through post-hoc interpretability methods that play a crucial role in fine-tuning the model before production.
Another critical question that we answered is model performance degradation in case of data shortage.
Our results suggest that knowledge injection gives the modeler more freedom in terms of the necessary data for proper model training.

This new methodology creates many new opportunities in terms of research.
One possible further research can be the investigation of more complex knowledge functions.
A second research path would be to enforce knowledge consistency within different scenarios, as in the case of time series.
These are only a few possible new research efforts that will pave the way for knowledge-informed DL.

\bibliographystyle{apalike}
\bibliography{biblio}

\appendix
\section{Dataset indicators}
\begin{landscape}
\begin{table}[]
  \centering
  \caption{Dataset financial indicators with corresponding description.}
\resizebox{\columnwidth}{!}{%
\begin{tabular}{@{}llll@{}}
\toprule
ID  & Description                                                                                                             & ID  & Description                                                                                   \\ \midrule
Attr 1  & net profit / total assets                                                                                               & Attr 33 & operating expenses / short-term liabilities                                                   \\
Attr 2  & total liabilities / total assets                                                                                        & Attr 34 & operating expenses / total liabilities                                                        \\
Attr 3  & working capital / total assets                                                                                          & Attr 35 & profit on sales / total assets                                                                \\
Attr 4  & current assets / short-term liabilities                                                                                 & Attr 36 & total sales / total assets                                                                    \\
Attr 5  & {[}(cash + short-term securities + receivables - short-term liabilities) / (operating expenses - depreciation){]} * 365 & Attr 37 & (current assets - inventories) / long-term liabilities                                        \\
Attr 6  & retained earnings / total assets                                                                                        & Attr 38 & constant capital / total assets                                                               \\
Attr 7  & EBIT / total assets                                                                                                     & Attr 39 & profit on sales / sales                                                                       \\
Attr 8  & book value of equity / total liabilities                                                                                & Attr 40 & (current assets - inventory - receivables) / short-term liabilities                           \\
Attr 9  & sales / total assets                                                                                                    & Attr 41 & total liabilities / ((profit on operating activities + depreciation) * (12/365))              \\
Attr 10 & equity / total assets                                                                                                   & Attr 42 & profit on operating activities / sales                                                        \\
Attr 11 & (gross profit + extraordinary items + financial expenses) / total assets                                                & Attr 43 & rotation receivables + inventory turnover in days                                             \\
Attr 12 & gross profit / short-term liabilities                                                                                   & Attr 44 & (receivables * 365) / sales                                                                   \\
Attr 13 & (gross profit + depreciation) / sales                                                                                   & Attr 45 & net profit / inventory                                                                        \\
Attr 14 & (gross profit + interest) / total assets                                                                                & Attr 46 & (current assets - inventory) / short-term liabilities                                         \\
Attr 15 & (total liabilities * 365) / (gross profit + depreciation)                                                               & Attr 47 & (inventory * 365) / cost of products sold                                                     \\
Attr 16 & (gross profit + depreciation) / total liabilities                                                                       & Attr 48 & EBITDA (profit on operating activities - depreciation) / total assets                         \\
Attr 17 & total assets / total liabilities                                                                                        & Attr 49 & EBITDA (profit on operating activities - depreciation) / sales                                \\
Attr 18 & gross profit / total assets                                                                                             & Attr 50 & current assets / total liabilities                                                            \\
Attr 19 & gross profit / sales                                                                                                    & Attr 51 & short-term liabilities / total assets                                                         \\
Attr 20 & (inventory * 365) / sales                                                                                               & Attr 52 & (short-term liabilities * 365) / cost of products sold)                                       \\
Attr 21 & sales (n) / sales (n-1)                                                                                                 & Attr 53 & equity / fixed assets                                                                         \\
Attr 22 & profit on operating activities / total assets                                                                           & Attr 54 & constant capital / fixed assets                                                               \\
Attr 23 & net profit / sales                                                                                                      & Attr 55 & working capital                                                                               \\
Attr 24 & gross profit (in 3 years) / total assets                                                                                & Attr 56 & (sales - cost of products sold) / sales                                                       \\
Attr 25 & (equity - share capital) / total assets                                                                                 & Attr 57 & (current assets - inventory - short-term liabilities) / (sales - gross profit - depreciation) \\
Attr 26 & (net profit + depreciation) / total liabilities                                                                         & Attr 58 & total costs /total sales                                                                      \\
Attr 27 & profit on operating activities / financial expenses                                                                     & Attr 59 & long-term liabilities / equity                                                                \\
Attr 28 & working capital / fixed assets                                                                                          & Attr 60 & sales / inventory                                                                             \\
Attr 29 & logarithm of total assets                                                                                               & Attr 61 & sales / receivables                                                                           \\
Attr 30 & (total liabilities - cash) / sales                                                                                      & Attr 62 & (short-term liabilities *365) / sales                                                         \\
Attr 31 & (gross profit + interest) / sales                                                                                       & Attr 63 & sales / short-term liabilities                                                                \\
Attr 32 & (current liabilities * 365) / cost of products sold                                                                     & Attr 64 & sales / fixed assets                                                                          \\ \bottomrule
\end{tabular}
}
\label{tab:app1}
\end{table}
\end{landscape}

\end{document}